# ENSEMBLE APPROACH FOR NATURAL LANGUAGE QUESTION ANSWERING PROBLEM

Anna Aniol
Department of Computer Science and Electrical Engineering
University of Science and Technology
Cracow, Poland
aa.annaaniol@gmail.com

Marcin Pietron
Department of Computer Science and Electrical Engineering
University of Science and Technology
Cracow, Poland
pietron@agh.edu.pl

## ABSTRACT

Machine comprehension, answering a question depending on a given context paragraph is a typical task of Natural Language Understanding. It requires to model complex dependencies existing between the question and the context paragraph. There are many neural network models attempting to solve the problem of question answering ([1], [3], [6], [12], [5]). The best models have been selected, studied and compared with each other. All the selected models are based on the neural attention mechanism concept. Additionally, studies on a SQUAD dataset were performed. The subsets of queries were extracted and then each model was analyzed how it deals with specific group of queries. Based on these three architectures ensemble model was created and tested on SQUAD dataset. It outperforms the best Mnemonic Reader model.

Keywords: Natural Language Processing, Machine comprehension, Deep learning

## 1 Introduction

Our main goal is to check how ensemble models can improve best machine comprehension architectures. At the first stage one of the the best three models were chosen for studying theirs accuracy in different types of queries. These are Bidirectional Attention Flow (BiDAF) [2], QANet [23] model and Mnemonic Reader [24]. All of them use deep learning models ([8], [9], [10]) combined with different types of attention mechanisms. Error analysis shows that each model obtains better results on different type of questions. Therefore, the models could be combined together in order to produce a better outcome for all questions of any kind. One of the most obvious approach is to build ensemble model. The main goal is to avoid weaknesses and use all strengths of analyzed architectures. The idea of building ensemble model is to combine predictions from different, well performing and separately trained models and calculate the actual prediction as the average or weighted predictions ([27], [28]). In presented work an answer comparison mechanism has been defined and implemented, to obtain a final answer based on separated answers given by chosen models. Before building the ensemble model comparative studies were performed between models, with particular reference to their attention layers and analysis of the results gained by models, including error analysis. The SQuAD dataset was used to train and evaluate the models. The proposed ensemble mechanism brings an improvement in predictions' accuracy.

## 2 Natural language question answering DL architectures

In this section the short overview of best question answering models is presented. The description mainly concentrated on main concepts, types of attention mechanisms and accuracy. All presented models have different types of attention mechanisms. Attention mechanism was present in many earlier models used in natural language tasks especially in question answering problem ([17], [16], [15], [4], [18], [19], [20], [14], [13]). The input data is transformed to glove representation ([11], [22]).

### 2.1 BIDAF

Bi-Directional Attention Flow network [2] is a hierarchical multi-stage architecture for a QA task. It allows for modeling the vector representation of the context paragraph at different structural levels: character-level, word-level and contextual-level. The architecture is based on the bidirectional attention flow mechanism ([16]). To avoid information loss caused by early summarization, attention is computed at each time step, instead of summarizing the context into one fixed-size vector. The result attended vector along with representations obtained in previous steps flows through to the subsequent modeling layer. The attention depends only on the query and the context at the current time step. It does not directly includes attention obtained in the previous steps. This is described by authors as memory-less attention and gives better results over dynamic attention. In order to obtain complementary information about the context and the query, the attention mechanism in BiDAF works in both ways - query-to-context and context-to-query attention is being computed.

### 2.2 Mnemonic Reader

The Reinforced Mnemonic Reader architecture [24] introduces two novel concepts to approaching the reading compre-hension task. First, authors present a re-attention mechanism. Second, they show a dynamic-critical reinforced learning approach to training models. With introducing the Reinforced Mnemonic Reader, authors address two issues of existing deep learning models for QA.

The first described issue is a problem affecting attention layers of existing models. A variety of successful neural attention mechanisms, such as bi-directional attention or co-attention have been proposed in a single-round alignment architecture. Multialigment architecture have been introduced in [25] in order to fully compose information of the inputs. It computes attentions repeatedly. That leads to attention redundancy, because the current attention can't be aware of what the previous attention attended to. At the same time, it leads to attention deficiency, as some parts of the input may never get attended by any of attentions. An answer to these problems is a re-attention mechanism, that memorizes past attentions. It uses the memorized results to refine the next attentions in a multi-round alignment architecture.

The second addressed issue is called a convergence suppression. It refers to a case of reinforcement learning which optimizes training towards F1 metric instead of EM using an estimated baseline in order to normalize the reward and reduce variances [26]. In such case, the convergence may get suppressed if the baseline is better than the reward, which is harmful if the inferior reward at some point overlaps with the ground truth. The normalized objective then discourages the prediction of ground truth positions. Authors proposed a dynamic-critical reinforcement learning, a novel approach to training. According to this approach, the reward and the baseline get dynamically decided according to random inference and greedy inference. Random inference and greedy inference are being considered two different sampling strategies. Random inference encourages exploration while greedy inference is meant for exploitation. The result that has a highest score becomes the reward, the other one becomes the baseline. It ensures that the normalized reward is positive. That eliminates the convergence suppression.

## 2.3 QANet

The main aim for introducing the QANet model was to create an architecture that delivers strong results on the SQuAD dataset, while being fast in training and inference. Models proposed before QANet achieve satisfying results but take a long time to train, as they use RNN component to process sequential inputs. In QANet, RNNs has been removed from the architecture and replaced by convolution and self-attention mechanisms in the encoder [19]. As per [23], QANet is 4.3 and 7.0 times faster than BiDAF in training and inference speed. Besides, it only needs one fifth of the training time to achieve BiDAF's top F1 score (77.3) on the dev set. Due to its training speed, QANet is a promising candidate to be scaled up to larger datasets in the future.

## 3 SQuAD dataset analysis

The section presents a statistical analysis of the SQuAD questions as well as an analysis of the results obtained by three standalone architectures with regard to the question classes. It shows how has the SQuAD data set been splitted in order to perform experiments. It presents the process of cross-referencing three standalone architectures, including assigning weights and resolving conflicts between the possible answers, in order to obtain the final prediction.

Fourteen question classes have been extracted based on thirteen types of question phrases (see Tables 1 and 2). If no question phrase matches the question, the question is labeled as undefined. Questions of the undefined type are 19:5% and 17:6% of the training and evaluation set respectively. Tables 1 and 2 show a detailed distribution of SQuAD questions from training and evaluation set over the defined question classes. Almost 53% of all the questions present in SQuAD have been defined as what questions. In the natural language, what questions are the most common, so the SQuAD dataset only reflects this trend. The only exception is the what time class, that has been considered as a separate type. Initially, there was also a separate class which (5% of the whole dataset). However, which and what can be used as an alternative to each other in most cases. Therefore, which and what classes are merged to what. Merging these two classes together has no influence on the final results.

| Type | Total count | Percentage |
|---|---|---|
| date | 770 | 0.9% |
| during | 1554 | 1.8% |
| how are | 180 | 0.2% |
| how big/size | 84 | 0.1% |
| how m/m | 5189 | 5.9% |
| how old | 104 | 0.1% |
| undefined | 17048 | 19.5% |
| what | 46114 | 52.6% |
| what time | 176 | 0.2% |
| when | 5431 | 6.2% |
| where | 884 | 1.0% |
| who | 8515 | 9.7% |
| whom | 354 | 0.4% |
| why | 1196 | 1.4% |
| SUM | 87599 | 100% |

Table 1: SQuAD training set - question classes breakdown

| Type | Total count | Percentage |
|---|---|---|
| date | 55 | 0.5% |
| during | 150 | 1.4% |
| how are | 33 | 0.3% |
| how big/size | 10 | 0.1% |
| how m/m | 681 | 6.4% |
| how old | 20 | 0.2% |
| undefined | 1856 | 17.6% |
| what | 5666 | 53.6% |
| what time | 14 | 0.1% |
| when | 692 | 6.5% |
| where | 95 | 0.9% |
| who | 1109 | 10.5% |
| whom | 39 | 0.4% |
| why | 150 | 1.4% |
| SUM | 10570 | 100% |

Table 2: SQuAD evaluation set - question classes breakdown

Performing the experiment required a dataset for evaluating the base models against different question classes, to obtain class-specific weights for the voting mechanism of the ensemble model. For this purpose, the original training set has been splitted into two parts: 95% of the training set remains the training data, and 5% of the training set becomes a pre-evaluation dataset. The base models are trained using the remaining 95% of data. Once the models are trained, they are evaluated against the pre-evaluation dataset. That allows to observe how accurate the models are for different question classes. Based on this observation, the ensemble model is constructed. Table 3 shows the distribution of the pre-evaluation questions over the question classes.

When deciding on how to construct the pre-evaluation dataset, two factors must have been taken info account. First, the more examples are in the pre-evaluation set, the more accurate the pre-evaluation process itself is. However, the larger the pre-evaluation set is, the smaller the remaining training set becomes. Splitting factor 0.05 (resulting in 95% remaining for training, 5% for pre-evaluation) is the smallest one that, when performing a random split, produces a pre-evaluation set where questions are distributed over classes similarly to the full SQuAD dataset distribution (see Table 3). At the same time, reducing the size of the training set by 5% does not weak the training process significantly. Splitting factor 0.01 and 0.1 have been tested and resulted in a worse general result.

| Type | Total count | Percentage |
|---|---|---|
| date | 31 | 0.6% |
| during | 60 | 1.2% |
| how are | 11 | 0.2% |
| how big/size | 4 | 0.1% |
| how m/m | 295 | 6.1% |
| how old | 6 | 0.1% |
| undefined | 811 | 16.8% |
| what | 2532 | 52.4% |
| what time | 11 | 0.2% |
| when | 390 | 8.1% |
| where | 67 | 1.4% |
| who | 502 | 10.4% |
| whom | 20 | 0.4% |
| why | 94 | 1.9% |
| SUM | 4834 | 100% |

Table 3: SQuAD pre-evaluation set - question classes breakdown

## 4 Ensemble approach

In the first step of creating ensemble architecture three base models have been trained on the full training set and evaluated against the evaluation data. The architectures have been implemented in PyTorch. The implementations are based on three existing repositories [29, 30], with some bug fixes, minor changes and performance tuning. Each training contains of 30 epochs and had the batch size set to 32.

Figures 1 and 2 and Table 4 present EM and F1 metrics gained by the standalone architectures. These results are the reference point for the results obtained by the ensemble model at a later stage. Data presented in Figures 1 and 2 confirm, that different models perform better for different types of questions. One globally best model does not have to be the best performing model for all types of questions. QANet achieves higher F1 for how are, what time and whom than Mnemonic Reader. BiDAF gains higher F1 for date-specific and how old questions. At the same time, Mnemonic Reader outperforms QANet and BiDAF in general.

| Model | F1 | EM |
|---|---|---|
| Mnemonic Reader | 81.57% | 73.25% |
| QANet | 79.43% | 69.84% |
| BiDAF | 75.17% | 64.12% |

Table 4: Results obtained by standalone models against the SQuAD evaluation dataset

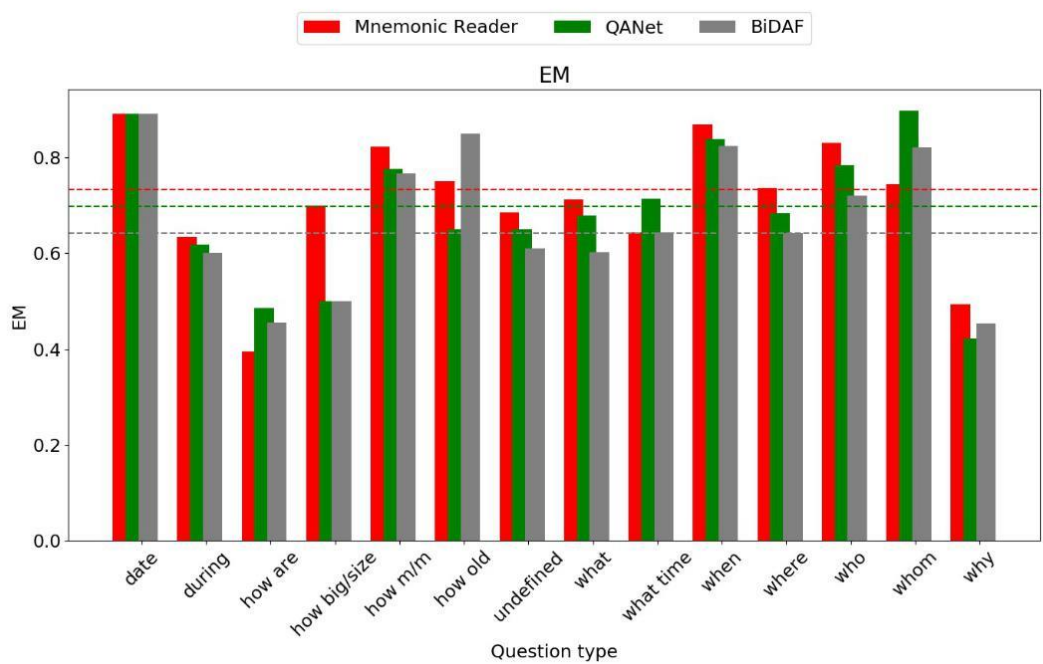

Figure 1: EM distribution over question classes (evaluation dataset)

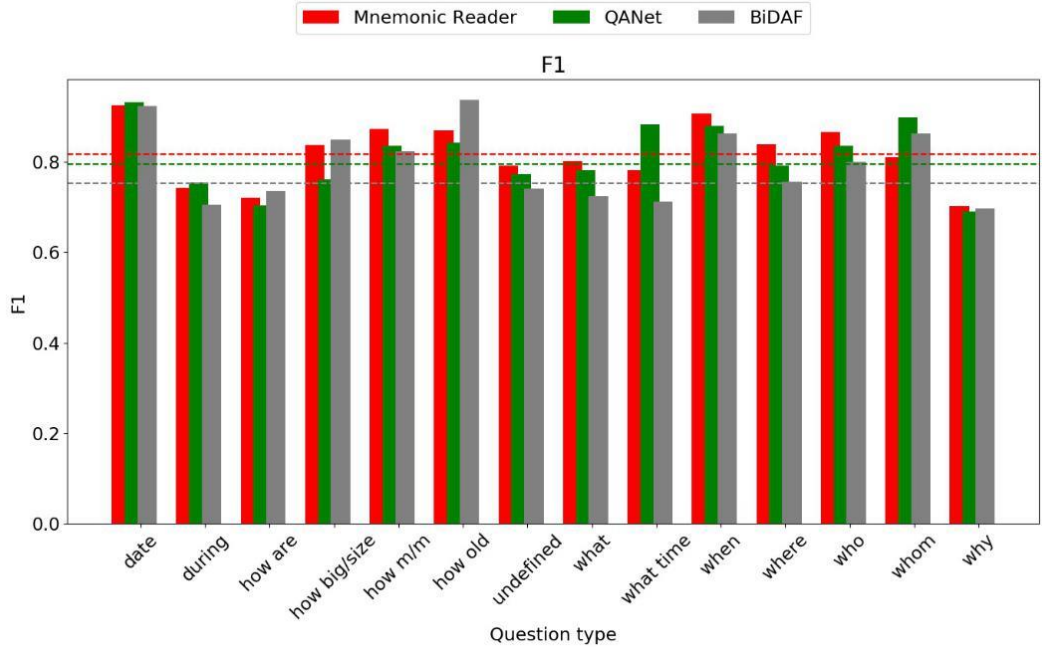

Figure 2: F1 distribution over question classes (evaluation dataset)

In order to verify, how similar to each other the predictions given by each pair of models are, the following analysis was performed. For each class of questions from the evaluation set the number of equal predictions (equal F1, equal EM per prediction) for each pair of models were calculated. The higher the similarity between all the predictions is, the smaller is the chance of the ensemble model to succeed.

Tables 5, 7 and 9 present the similarities between predictions obtained by each pair of models on the evaluation set. At the same time, tables 6, 8 and 10 show what is an average F1 an how many times EM is True among the predictions that are equal between two models.

| Type | Equal F1 | Equal EM | Total count |
|---|---|---|---|
| date | 51 (92.7%) | 51 (92.7%) | 55 |
| during | 113 (75.3%) | 121 (80.7%) | 150 |
| how are | 21 (63.6%) | 30 (90.9%) | 33 |
| how big/size | 7 (70.0%) | 8 (80.0%) | 10 |
| how m/m | 540 (79.3%) | 550 (80.8%) | 681 |
| how old | 17 (85.0%) | 18 (90.0%) | 20 |
| undefined | 1342 (72.3%) | 1512 (81.5%) | 1856 |
| what | 4059 (71.6%) | 4489 (79.2%) | 5666 |
| what time | 9 (64.3%) | 11 (78.6%) | 14 |
| when | 603 (87.1%) | 623 (90.0%) | 692 |
| where | 73 (76.8%) | 76 (80.0%) | 95 |
| who | 914 (82.4%) | 935 (84.3%) | 1109 |
| whom | 32 (82.1%) | 33 (84.6%) | 39 |
| why | 71 (47.3%) | 108 (72.0%) | 150 |
| SUM | 7852 (74.3%) | 8565 (81.0%) | 10570 |

Table 5: Mnemonic Reader (10570) vs. QANet (10482)

| Mean of equal F1s | Number of Trues in equal EMs |
|---|---|
| 87.6% | 6540 (76.4%) |

Table 6: Mnemonic Reader vs. QANet - mean equal metrics

| Type | Equal F1 | Equal EM | Total count |
|---|---|---|---|
| date | 51 (92.7%) | 51 (92.7%) | 55 |
| during | 102 (68.0%) | 113 (75.3%) | 150 |
| how are | 23 (69.7%) | 31 (93.9%) | 33 |
| how big/size | 7 (70.0%) | 8 (80.0%) | 10 |
| how m/m | 552 (81.1%) | 565 (83.0%) | 681 |
| how old | 18 (90.0%) | 18 (90.0%) | 20 |
| undefined | 1286 (69.3%) | 1473 (79.4%) | 1856 |
| what | 3723 (65.7%) | 4248 (75.0%) | 5666 |
| what time | 10 (71.4%) | 12 (85.7%) | 14 |
| when | 602 (87.0%) | 625 (90.3%) | 692 |
| where | 71 (74.7%) | 78 (82.1%) | 95 |
| who | 840 (75.7%) | 868 (78.3%) | 1109 |
| whom | 30 (76.9%) | 32 (82.1%) | 39 |
| why | 76 (50.7%) | 114 (76.0%) | 150 |
| SUM | 7391 (69.9%) | 8236 (77.9%) | 10570 |

Table 7: Mnemonic Reader (10570) vs. BiDAF (10570)

| Mean of equal F1s | Number of Trues in equal EMs |
|---|---|
| 86.7% | 6093 (74.0%) |

Table 8: Mnemonic Reader vs. BiDAF - mean equal metrics

| Type | Equal F1 | Equal EM | Total count |
|---|---|---|---|
| date | 52 (94.5%) | 53 (96.4%) | 55 |
| during | 94 (62.7%) | 114 (76.0%) | 150 |
| how are | 23 (69.7%) | 32 (97.0%) | 33 |
| how big/size | 7 (70.0%) | 8 (80.0%) | 10 |
| how m/m | 543 (79.7%) | 561 (82.4%) | 681 |
| how old | 16 (80.0%) | 16 (80.0%) | 20 |
| undefined | 1231 (66.3%) | 1451 (78.2%) | 1856 |
| what | 3586 (63.3%) | 4193 (74.0%) | 5666 |
| what time | 8 (57.1%) | 11 (78.6%) | 14 |
| when | 591 (85.4%) | 618 (89.3%) | 692 |
| where | 72 (75.8%) | 83 (87.4%) | 95 |
| who | 831 (74.9%) | 876 (79.0%) | 1109 |
| whom | 34 (87.2%) | 34 (87.2%) | 39 |
| Why | 65 (43.3%) | 106 (70.7%) | 150 |
| SUM | 7153 (67.7%) | 8156 (77.2%) | 10570 |

Table 9: QANet (10482) vs. BiDAF (10570)

| Mean of equal F1s | Number of Trues in equal EMs |
|---|---|
| 86.5% | 5858 (71.8%) |

Table 10: QANet vs. BiDAF - mean equal metrics

The numbers presented in Tables 5 to 10 show, that predictions obtained by the base models are diverse in general as well as within each question class. That observation confirms that the ensemble model based on the class-specific pre-evaluation may lead to improving results of the best performing model.

Presented ensemble model is based on a class-specific weighted voting. Algorithm 1 presents, how the class-specific weights are obtained. It is an average F1 metric obtained after evaluation the model on all questions from specific class. Algorithm 2 describes, how does the voting mechanism work. It gathers all candidate answers from all models and if at least two models give duplicate answer their weights are added. Finally, the candidate answer is this one with highest weight or returned by globally best model (in specific class) when no duplicated answer exists.

---

**Algorithm 1** Obtaining class-specific voting weights
---
    Training data: 95% of the SQuAD training set
    Pre-evaluation data: 5% of the SQuAD training set
1: Each model is trained on the Training data.
2: Each model from step 1. is pre-evaluated against the pre-evaluation data.
3: Class-specific weights for each model are obtained. The class-specific weight is an average F1 obtained by the model when pre-evaluating on all the questions of that class.

---

**Algorithm 2** Weighted class-specific voting ensemble
---
    Training data: full SQuAD training data set
    Evaluation data: full SQuAD evaluation data set
1: Each model is trained on the Training data.
2: For each evaluation question, a question class is identified.
3: Each trained base model is asked to answer the questions from the Evaluation data, resulting in three candidate answers per question.
4: Each candidate answer gets a weight that corresponds to the question class and the answering model.
5: if class is different from undefined then
6:    if one candidate answer is a duplicate of another one then
7:       Their weights are added together
8:       The ensemble returns the candidate answer that has the highest weight.
9:    else
10:       The candidate answer produced by the globally best model is returned
11: else
12:    The candidate answer produced by the globally best model is returned

---

For the purpose of verifying, the algorithm that ignores question classes was set up (see Algorithm 3 and 4).

**Algorithm 3** Obtaining voting weights (no classes)
  Training data: 95% of the SQuAD training set
  Pre-evaluation data: 5% of the SQuAD training set
1: Each model is trained on the Training data.
2: Each model from step 1. is pre-evaluated against the pre-evaluation data.
3: General weights for each model are obtained. Each weight is an average F1 obtained by the model during pre-evaluation.

**Algorithm 4** Weighted voting ensemble (no classes)
  Training data: full SQuAD training data set
  Evaluation data: full SQuAD evaluation data set
1: Each model is trained on the Training data.
2: Each trained base model is asked to answer the questions from the Evaluation data, resulting in three candidate answers per question.
3: Each candidate answer gets a weight obtained in Algorithm 3.
4: if one candidate answer is a duplicate of another one then
5:     their weights are added together
6: The ensemble returns the candidate answer that has the highest weight.

## 5 Results

Table 11 shows the F1 and EM metrics achieved by the ensemble model when testing against the SQuAD evaluation data set. The class-aware ensemble model outperforms Mnemonic reader by 0.39 percentage point in F1 score and by 0.52 percentage point in EM score. At the same time, it outperforms the non class-aware ensemble in F1 and EM by 0.1 percentage point and 0.05 percentage point respectively. As expected, it resulted in worse ensemble results than the class-aware ensemble achieves.

| Model | F1 | EM |
|---|---|---|
| Mnemonic Reader | 81.57% | 73.25% |
| QANet | 79.43% | 69.84% |
| BiDAF | 75.17% | 64.12% |
| Class-aware ensemble | 81.96% | 73.77% |
| Non class-aware ensemble | 81.86% | 73.72% |

Table 11: Results of the ensemble model

In order to demonstrate where does the improvement come from, the ensemble results over classes must be presented. Figures 3 and 4 show results of the pre-evaluation.

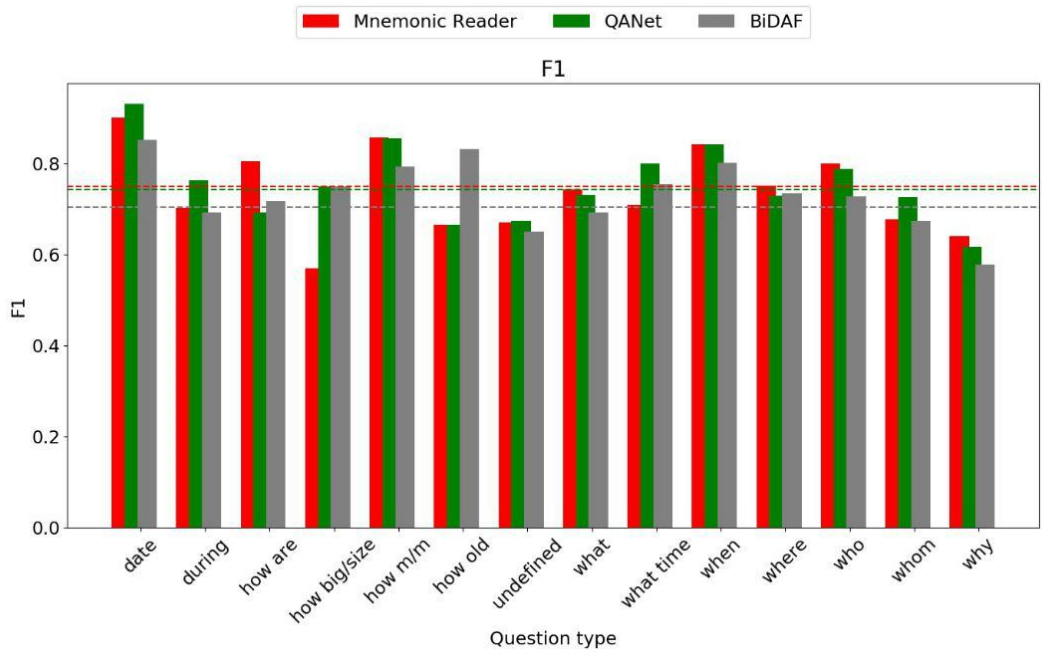

Figure 3: F1 distribution over question classes (pre-evaluation data)

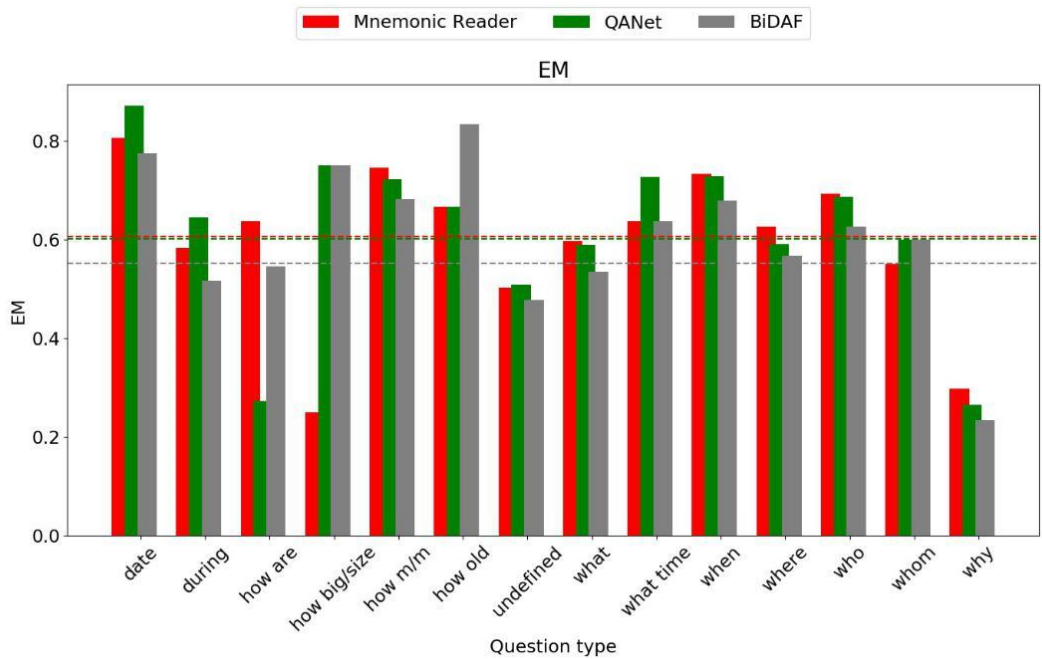

Figure 4: EM distribution over question classes (pre-evaluation data)

Based on the results presented above, the voting weights are obtained. For example, for date-specific questions QANet's answers have the highest weight, as the average F1 of QANet's pre-evaluation answers is higher than Mnemonic's and BiDAF's answers. Due to this fact, for date-specific questions in the evaluation set the ensemble returns the answer produced by QANet (as long as there are no duplicated candidate answers). Figures 5 ans 6 present the detailed ensemble results over question classes.

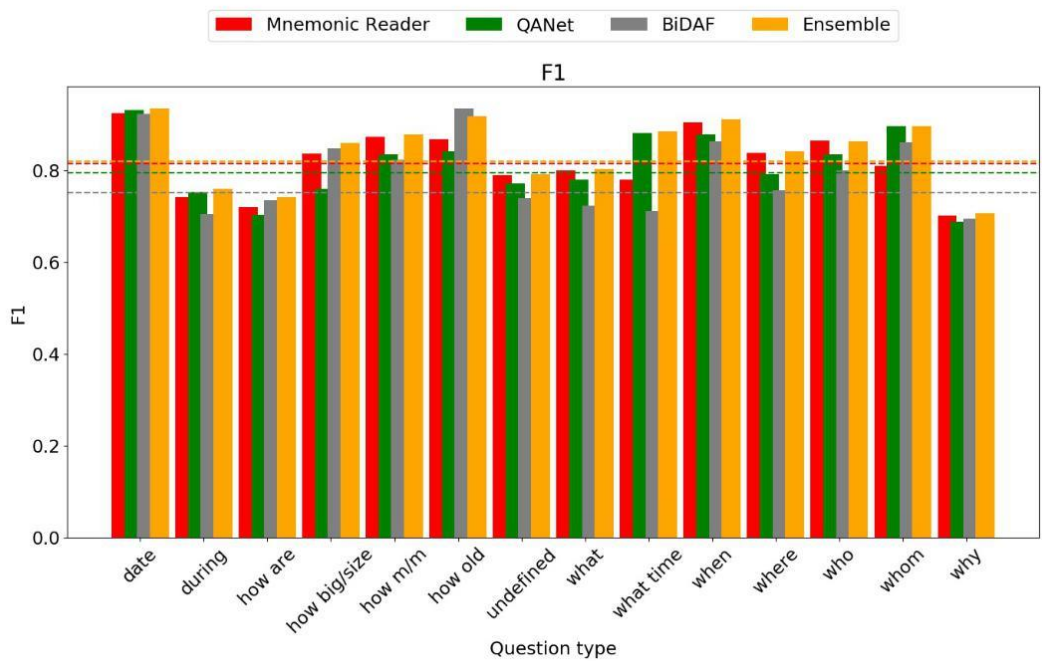

Figure 5: F1 Results over question classes (evaluation data)

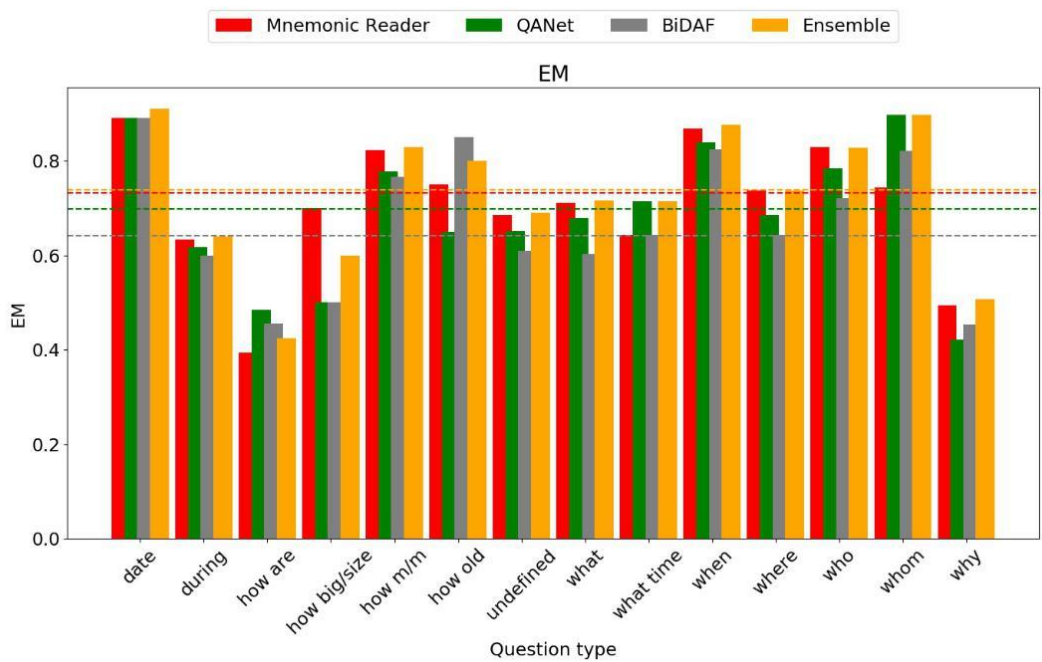

Figure 6: EM Results over question classes (evaluation data)

Some interesting observations can be made here. For example, for date-specific question, the ensemble model achieves higher F1 and EM than the best performing standalone model (QANet). The reason is, that there are some date-specific questions in the evaluation data, for which QANet returns an incorrect prediction, while both Mnemonic Reader and BiDAF return the correct one. In this case, weights of Mnemonic and BiDAF are added together and result in a higher weight than the QANet's weight. Another interesting behaviour can be noted for the how old questions. The ensemble model performs here worse than BiDAF, that has the highest voting weight. However, when BiDAF returns the correct answer while Mnemonic and QANet return an incorrect but equal prediction, their weights are added together and the incorrect prediction is returned. In case of the how old questions it would be better not to add the weights of equal predictions together, but to only take the maximum of them. However, this approach was tested globally in all classes but adding weights of equal answers together gives better results.

Next tested approach was modifications of Algorithm 2 with no if-else block to distinguish between undefined questions and other question classes. It resulted in a different ensemble behaviour for the undefined questions. The Undefined class requires weighted voting as other classes did. It had a negative impact on the ensemble results. Another possible modification is to use average EM instead of average F1 to obtain class weights in Step 2. of Algorithm 1. However, it results in worse ensemble results, both in F1 and EM score.

In order to test, if any other classification criteria can perform better for the voting ensemble, we have tested some alternative classification methods. We performed questions classification based on their length, where length is the number of words in a question. It doesn't matter how dense the criteria is, it brings no improvement. No relation between the question's length and the model's performance has been demonstrated in here. It resulted in the ensemble model achieving the same results as the best performing standalone model.

## 6 Conclusions

The presented results shows that ensemble algorithm based on deep learning models with different architectures can improve the results in natural language question answering problem. The deeper analysis of input data was required to achieve better results than best deep learning model. The further improvements will concentrate on further analysis of models and build one that be able to achieve comparable results to ensemble on using combinations of techniques used in presented architectures or its modifications.